\documentclass[10pt,twocolumn,letterpaper]{article}

 \usepackage{cvpr}              

\usepackage[dvipsnames,table]{xcolor}

\definecolor{white}{rgb}{1,1,1}
\definecolor{lightbluishgrey}{rgb}{0.76471,0.84824,0.91647}
\definecolor{derekTableBlue}{rgb}{0.565,0.847,0.769} 
\definecolor{cvprblue}{rgb}{0.21,0.49,0.74}
\newcommand{\revision}[1]{#1}

\usepackage{layouts}

\usepackage[mathletters]{ucs}
\usepackage[utf8x]{inputenc}

\usepackage{tabularx}
\usepackage{booktabs}
\usepackage{makecell}

\usepackage{subcaption}
\usepackage{wrapfig}
\usepackage{graphicx}
\usepackage{caption}
\usepackage{float}

\usepackage{listings}

\usepackage{booktabs} 
\usepackage{blindtext}
\usepackage{caption}
\usepackage{layouts}
\usepackage{amsmath}
\usepackage{bbm}
\usepackage{physics}
\usepackage{float}
\usepackage{multicol}  

\usepackage[pagebackref,breaklinks,colorlinks,citecolor=cvprblue]{hyperref}

\def\paperID{376} 
\def\confName{3DV\xspace}
\def\confYear{2025\xspace}

\title{Rigid Body Adversarial Attacks}

\author{Aravind Ramakrishnan\\
University of Toronto\\
{\tt\small aravind@cs.toronto.edu}
\and
David I.W. Levin\\
University of Toronto\\
NVIDIA\\
{\tt\small diwlevin@cs.toronto.edu}
\and
Alec Jacobson\\
University of Toronto\\
Adobe Research\\
{\tt\small jacobson@cs.toronto.edu}
}

\begin{document}

\twocolumn[{%
\renewcommand\twocolumn[1][]{#1}%
\maketitle
 \vspace{-1.5em}
\includegraphics[width=\linewidth]{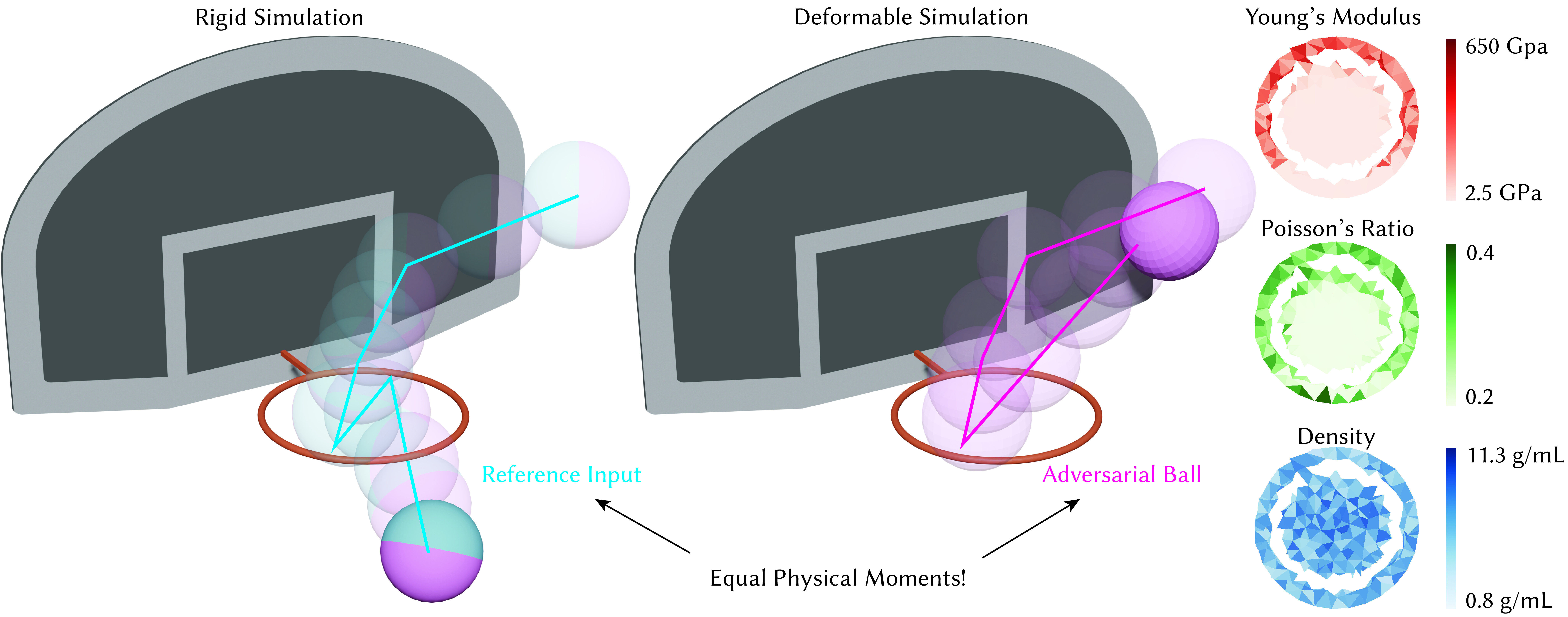}
\captionof{figure}{We construct an adversarial ball (right) out of perceptually stiff materials (with a minimum Young's modulus of 2.5 GPa), such that it results in a maximally different deformable simulation trajectory (middle) compared to a reference ball while having identical physical moments and thus trajectories in a rigid body simulation (left).}
 \label{fig:teaser}
 \vspace{0.5em}
}]

\begin{abstract}
Due to their performance and simplicity, rigid body simulators are often used in applications where the objects of interest can considered very stiff. However, no material has infinite stiffness, which means there are potentially cases where the non-zero compliance of the seemingly rigid object can cause a significant difference between its trajectories when simulated in a rigid body or deformable simulator.

Similarly to how adversarial attacks are developed against image classifiers, we propose an adversarial attack against rigid body simulators. In this adversarial attack, we solve an optimization problem to construct perceptually rigid adversarial objects that have the same collision geometry and moments of mass to a reference object, so that they behave identically in rigid body simulations but maximally different in more accurate deformable simulations. We demonstrate the validity of our method by comparing simulations of several examples in commercially available simulators.
\end{abstract}    
\section{Introduction}\label{sec:intro}

When simulating stiff objects, rigid body simulators are extremely popular and are often used in lieu of more accurate deformable simulators due to their performance and simplicity. 
Of course, even very stiff materials are deformable at some scale. These deformations change how objects respond to contacts and how internal stresses propagate through an object. Moreover, spatial variations of material properties in an object are not captured by the object-level parameters of a rigid body simulator: mass, center of mass, and moment of inertia. 

\revision{As a result, it is possible that simulating certain objects with rigid body methods can lead to less accurate results. This can be dangerous in contexts where safety of the system is determined by the robustness of the underlying model.}
For example, in machine learning, there is increasing interest in learning physical models (e.g. \cite{physml1,physml2,physml3,physml4}), including rigid body dynamics (e.g. \cite{rbdml1,rbdml2,rbdml3}), for use in various downstream applications where inaccuracies can cause problems.
\revision{In graphics, if a rigid body simulation might be used for previsualization and replaced with an elasticity simulation later, abrupt changes could lead to surprises.} 
In robotics, Sim2Real training (see \cite{sim2real1,sim2real2,sim2real3}) based on vulnerable physics simulation could be dangerous in safety critical tasks.

In deep learning, such vulnerabilities are studied in the form of \emph{adversarial attacks}, where seemingly normal malicious inputs to the network are generated (typically as imperceptible perturbations to a reference input) in order to cause the model to make mistakes. For instance, images with some underlying change such as small modifications to pixel values or lighting can be used to trick image classifiers into misclassifying them, which can have grave consequences if the classifier was used in a safety critical application \cite{chakraborty2021survey, xu2020adversarial}.
These attacks exist in physical real-world applications. A few examples include the use of adversarial background music to disrupt the functionality of the voice assistant tool Amazon Alexa \cite{adv-music}, the use of makeup to cause face recognition software to fail \cite{adv-makeup}, and even placing small markings on the road to trick Tesla vehicles' autopilot lane detection software \cite{keen-lab-tesla}. 
In machine learning, researching these adversarial attacks was the first step towards improving safety, eventually leading to adversarial training which turned attacks into strategies to make models more robust.

In this paper, we propose using optimization techniques to construct adversarial objects using physically reasonable materials, which will behave identically to a reference object in rigid body simulation, but maximally different in more physically accurate deformable simulation. As rigid body simulators use only the collision geometry and the moments of the object, the adversarial objects require identical external geometry and first three moments of mass to the reference object so that they are indistinguishable in the rigid body setting.
To achieve this objective, we first define a cost function that encodes the difference in the simulation result of the reference and adversarial object. We choose the degrees of freedom of the optimization to be the object's material distribution and internal geometry. By using the adjoint method we can efficiently compute gradients of the cost with respect to these degrees of freedom, allowing the use of a descent method to determine the internal geometry and materials that minimizes the cost subject to having certain first three moments of mass. 

We demonstrate the efficacy of our method by constructing several adversarial objects and comparing the results of their simulations with their reference in \textsc{Polyfem}, a commercially available simulator. Thus, we show that robotics planning and control tools make a potentially dangerous model assumption when using rigid body simulators.
\section{Related Works}\label{sec:Related Works}
To construct adversarial objects, our work draws on ideas from inverse problems, machine learning, and simulation.

\subsection{Inverse Problems}
Consider a system whose behavior is governed by some set of parameters. In a \textit{forward problem}, the goal is to compute the evolution of that system given its parameters. In the corresponding \textit{inverse problem}, the goal is to determine the parameters of the system from observations of the evolution of the system - either from physical measurements or via simulation (see e.g. \cite{inverse-problems-book1, inverse-problems-book2}). 
\revision{In machine learning and vision, common inverse problems include inverse kinematics (e.g. \cite{invkin2, invkin3}) and imaging (e.g. \cite{invimage2, invimage3}), though there have been works that tackle more broad inverse problems in physics using invertible neural networks \cite{inverse-phys-inn} and approximate inverse simulation with a learned correction function \cite{inverse-phys-score-matching}}.

In the context of this work, the system we consider is the elastodynamics simulation of an object, and our inverse problem to determine its material parameters and internal geometry.
In mechanics, \citet{XU2023115852} use physics informed neural networks to solve inverse problems with linear elastic and hyperelastic materials where they determine loads and internal pressures.
In computer animation, there has been some work in solving inverse problems to optimize the trajectories of objects. Approaches include determining control forces via sequential quadratic programming \cite{spacetime-constraints}, determining physical parameters (i.e. initial positions and velocities, surface normal variations, etc.) that minimize an energy subject to position constraints of an object at chosen times \cite{popovic2000interactive}, and simulating backwards using time-reversed simulators to determine initial conditions \cite{backward-rigid-body}. Unlike these prior works, we are optimizing for material properties rather than loads, control forces, or initial conditions.

\begin{figure}[t!]
	\centering
	\includegraphics[width=\linewidth]{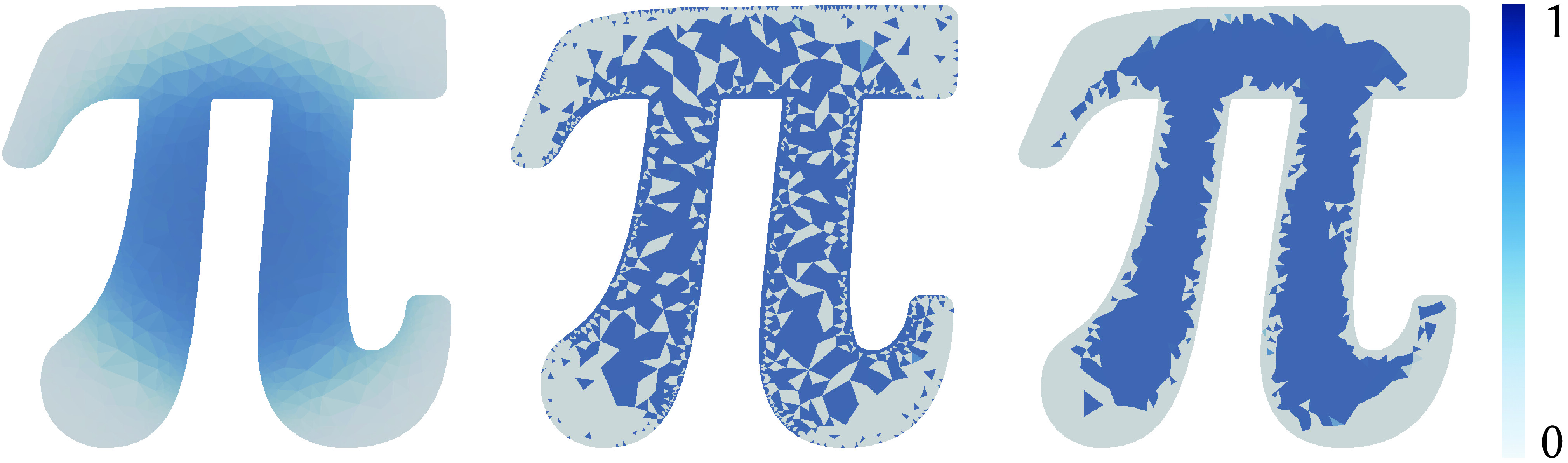}
	\vspace{-0.7cm}
	\caption{For a given geometry, there can be many mass density distributions that have identical moments of mass. Above are three visually different density distributions on a mesh that have the same total mass, center of mass, and moment of inertia.}
	\label{fig:density-shapes}
\end{figure}

Optimizing the internal geometry of our adversarial object leads us to the topology optimization inverse problem, where one finds the optimal allocation of a fixed amount of material over a design domain such that the compliance of the design is minimized when subjected to a load. There are a few broad approaches to topology optimization including nodal/element based design variables \cite{XIE1993885,simp-bendsoe}, level sets \cite{ALLAIRE20021125, WANG2003227}, and implicit neural representations \cite{ntopo}.
In our work, we follow the solid isotropic material with penalization parametrization (SIMP) approach of defining a per element occupancy and using a penalizing power-law to scale stiffness (see \cite{bendsoe-book}).
However, unlike the standard topology optimization problem, we are not optimizing for minimum compliance, do not have a fixed amount of material, have a continuous spread of allowed materials rather than one or a few, and furthermore, we are interested in optimizing over dynamic loads. 
Instead of a fixed amount of material, our constraint is matching moments of mass. This is a very similar problem to \citet{same-stats}, where the authors construct datasets of 2D points such that they have the same mean (first moment), standard deviation (second moment about the mean), and Pearson's R value to those of some reference dataset. By using a simulated annealing optimization scheme, they can direct the new dataset to be along a target shape. We instead leave our material density distribution as an untargeted optimization, where it may take any values such that the moments match (see Fig. \ref{fig:density-shapes}).

\subsection{Adversarial Attacks}
Our method is inspired by adversarial attacks in machine learning, where an adversarial input is constructed by causing small perturbations to a reference.
There has been considerable interest for adversarial attacks in various contexts and applications, including images \cite{Szegedy2014, goodfellow2014explaining, moosavi2016deepfool,kurakin2018adversarial, parametric-adversary-derek}, videos \cite{Li2018AdversarialPA,sparse-adv-video, linxi-video,Li2021AdversarialAO,10378215-kim}, audio \cite{smack-audio,Chen2019WhoIR,aaecaptcha,Liu2019WeightedSamplingAA}, \revision{geometric representations of 3D shapes \cite{lang2021geometric-adv, hu2023geometric-adv, Stolik2022SAGASA}}, 3D printed objects that fool classifiers \cite{pmlr-v80-athalye18b, Huang2023TowardsTT}, machine learning components of industrial control systems \cite{aa-erba, ANTHI2021102717, figueroa-ic}, and even control signals in robots \cite{otomo2022adversarial}.

Most relevant to our work is the intersection of physical inverse design problems and adversarial attacks. To our knowledge, the only study combining the two fields is \citet{azakami2022adversarial}, who perform an attack over the shape of legged robots that are simulated in MuJoCo. These robots have body components parameterized by their lengths and thicknesses. The authors use a differential evolution optimization scheme to find the parameters that are close to those of the original robot but induce a failure in the control, causing them to fall over. However, the parameter changes may cause the new design to have different centers of mass and broken symmetries.

In machine learning, adversarial attacks have led to adversarial training, where model robustness is improved by training on adversarial examples \cite{ijcai2021p591, QIAN2022108889}. We think our adversarial rigid objects can similarly be used for improved validation and training of learned physics simulation and robotics planning and control methods.

\subsection{Simulation}
Both rigid body (RBD) and deformable simulation techniques are popular in robotics \cite{liu2021role}.

\subsubsection{Rigid Body Simulation}
\revision{
Dynamic simulation of highly stiff objects is a common task across several fields. A rigid body is an idealized model, where the simulated object is treated as infinitely stiff - external stimulus like forces or impulses are propagated almost instantaneously across the object eliminating relative deformations \cite{affine-body-dynamics}. As a result, rigid body simulators require only per object rotational and translational degrees of freedom, rather than requiring degrees of freedom for each material point in the object to capture deformations. This multiple orders of magnitude reduction in degrees of freedom per object makes the rigid body model extremely performant and useful in a variety of interactive applications. Its speed also means that rigid body simulators are often used for simulating highly stiff objects in robotics and are used in robotics control \cite{posa-control, rigid-control-2, rigid-soft-robot}. 
}

\revision{Rigid body simulation has been used for modelling multi object scenes with collision since the late 1980's \cite{baraff1989rbd, hahn1988realistic, moore1988collision}, and a comprehensive survey on the topic can be found in \citet{rbd-survey}.}
\revision{The most involved process in rigid body simulation is in the handling of collision and contact.} Two main classes of rigid body simulators have emerged, some handling contact using penalties or barriers (e.g. \cite{RigidIPC, drumwright-rbd}), and others using constraints \revision{encoding non-penetration} formulated as linear complementarity or velocity-level approximate linear complementarity problems (e.g. \cite{erleben2007velocity, trinkle-rbd}).

\subsubsection{Deformable Simulation}
The simulation of deformable bodies has been extensively studied both in robotics and in the broader academic community \revision{through various methods including mass-spring systems, the finite element method (FEM), the boundary element method, and the material point method.}
We choose to use FEM for our deformable simulation, which is used to solve boundary value problems over potentially intricate domains by using a discretization into elements (in our case tetrahedra). For a detailed introduction, see \cite{fem-book, dynamic-deformables, sifakis2012fem}. In FEM, one can choose between different constitutive models for materials depending on the application - we choose the stable Neohookean energy \cite{dynamic-deformables, smith-kim-nh}. While FEM can be very accurate and is constantly being improved, it may still be too slow to be used in certain applications.

\subsubsection{Differentiable Simulation}
In order to solve the inverse problem using a descent method, we need to be able to calculate gradients through the simulation. Differentiable simulation is a powerful tool for physics based learning and control problems, and has picked up in prevalence both in the context of rigid bodies \cite{neuralsim, JADE, BelbutePeres2018EndtoEndDP, diffsdfsim} and deformable objects \cite{gradsim, du2021diffpd, qiao2021differentiable, gjoka2022differentiable, add-diff-phys, diffcloth}. We follow the approach taken by \citet{gradsim}, where we implement our physics simulator with an autodifferentiation framework to which we provide manual derivatives at each timestep.

\section{Adversarial Objects}\label{sec:method}
Given a solid reference object $\Omega \subset \mathbb{R}^3$ and its moments of mass $m_0 \in \mathbb{R}_{>0}$, $m_1 \in \mathbb{R}^3$, and $m_2 \in \mathbb{R}^{3x3}$, we optimize a material distribution (consisting of Young's modulus $Y$, Poisson's ratio $\nu$, mass density $\rho$, and material occupancy $\alpha$) to maximally change the trajectories after simulation:
\begin{gather}
\underset{Y,\ \nu,\ \rho,\ \alpha}{\mathrm{argmax}} \quad  \int_\Omega || q_\text{adv}(t_\text{end}) -  q_\text{ref}(t_\text{end}) ||^2 dV  \label{eq:continuous} \\
\text{s.t.}\quad                    Y(x) \in [Y_\text{min}, Y_\text{max}]\quad \forall x \in \Omega  \tag{\ref{eq:continuous}a} \\
\phantom{\text{s.t.}}\quad   \nu(x) \in [\nu_\text{min}, \nu_\text{max}] \quad \forall x \in \Omega   \tag{\ref{eq:continuous}b} \\
\phantom{\text{s.t.}}\quad   \rho(x) \in [\rho_\text{min}, \rho_\text{max}] \quad \forall x \in \Omega  \tag{\ref{eq:continuous}c} \\
\phantom{\text{s.t.}}\quad   \alpha(x) \in \{0, 1\} \quad \forall x \in \Omega\strut^\mathrm{o} \tag{\ref{eq:continuous}d} \\
\phantom{\text{s.t.}}\quad   \alpha(x) = 1 \quad \forall x \in \partial \Omega  \tag{\ref{eq:continuous}e} \\
\phantom{\text{s.t.}}\quad   \int_\Omega \rho(x)\alpha(x) dV = m_0 \tag{\ref{eq:continuous}f} \\
\phantom{\text{s.t.}}\quad   \int_\Omega x \rho(x)\alpha(x) dV = m_1 \tag{\ref{eq:continuous}g} \\
\phantom{\text{s.t.}}\quad   \int_\Omega ((x \cdot x)I - xx^T) \rho(x)\alpha(x) dV = m_2, \tag{\ref{eq:continuous}h}
\end{gather}
where $q_\text{ref}$ is the simulation result of the reference object, $q_\text{adv}$ is the simulation result of the adversarial object, $q(t_\text{end})$ represents the final state of the simulation. \revision{We delegate detailed discussion the physical meaning of material parameters (e.g. \citet{materials-strength-book})}. Looking at the constraints, Eq. \ref{eq:continuous}(a-c) ensure that material parameters stay within a specified range of perceptually stiff materials. Eq. \ref{eq:continuous}(d-e) allow changing the internal geometry of the object while ensuring that the boundary of the domain remains the same (same collision geometry). And Eq. \ref{eq:continuous}(f-h) ensure that the moments of the object are the same (same behavior if reduced to RBD).
\revision{
Together, the constraints enforce our notion of ``imperceptible perturbation" to the rigid body simulator - the object is made of only highly stiff materials (such that one may reasonably choose to simulate with rigid body techniques - see Supplemental D), the object has the same external geometry to the reference (so that it experiences identical contacts), and it has identical moments of mass (meaning identical response to the reference in rigid body simulation due to the reduced degrees of freedom).
}

Typically, a rigid body simulator takes as input the boundary of the domain $\partial \Omega$ as, e.g., a triangle mesh and the moments as parameters. We choose to discretize our problem by tetrahedralizing the interior of the input mesh and treating $Y$, $\nu$, $\rho$, and $\alpha$ as per-tetrahedron quantities:
\begin{gather}
\underset{Y,\ \nu,\ \rho,\ \alpha}{\mathrm{argmax}} \quad  \sum_i V_i || q_i^\text{adv}(t_\text{end}) -  q_i^\text{ref}(t_\text{end}) ||^2  \label{eq:discrete} \\
\text{s.t.}\quad                    Y_i \in [Y_\text{min}, Y_\text{max}]\quad \forall i  \tag{\ref{eq:discrete}a} \\
\phantom{\text{s.t.}}\quad   \nu_i \in [\nu_\text{min}, \nu_\text{max}] \quad \forall i   \tag{\ref{eq:discrete}b} \\
\phantom{\text{s.t.}}\quad   \rho_i \in [\rho_\text{min}, \rho_\text{max}] \quad \forall i  \tag{\ref{eq:discrete}c} \\
\phantom{\text{s.t.}}\quad   \alpha_i \in \{0, 1\} \quad \forall i \in \Omega\strut^\mathrm{o} \tag{\ref{eq:discrete}d} \\
\phantom{\text{s.t.}}\quad   \alpha_i = 1 \quad \forall i \in \partial \Omega  \tag{\ref{eq:discrete}e} \\
\phantom{\text{s.t.}}\quad   \sum_i V_i \rho_i \alpha_i = m_0 \tag{\ref{eq:discrete}f} \\
\phantom{\text{s.t.}}\quad   \sum_i \rho_i \alpha_i \int_i x dV = m_1 \tag{\ref{eq:discrete}g} \\
\phantom{\text{s.t.}}\quad   \sum_i \rho_i \alpha_i \int_i ((x \cdot x)I - xx^T) dV = m_2, \tag{\ref{eq:discrete}h}
\end{gather}
where $i$ refers to an individual tetrahedron in $\Omega$ (now represented with a tetrahedral mesh: $V$, $T$).
\begin{figure*}[t]
 \centering
 \includegraphics[width=\linewidth]{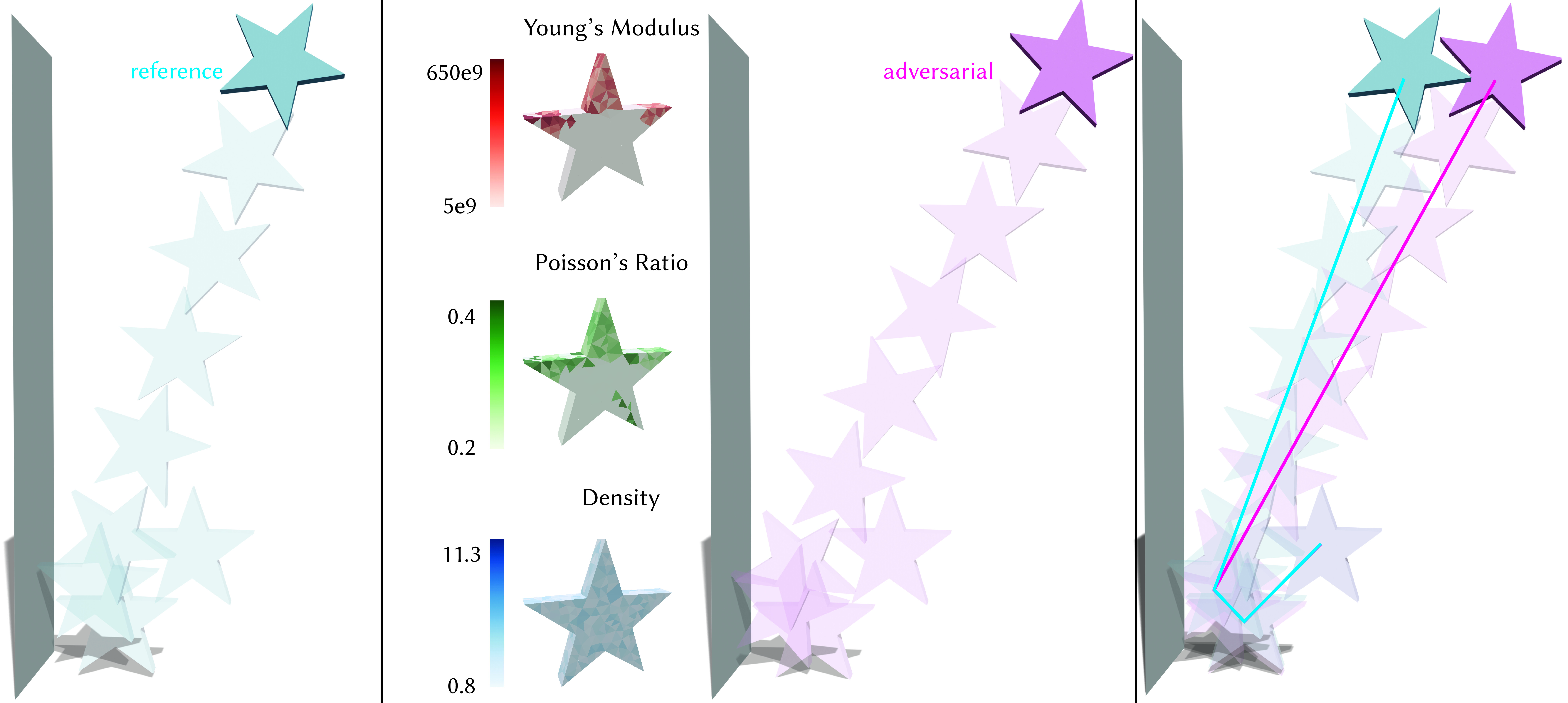}
 \vspace{-0.7cm}
 \caption{We simulate a star colliding off of the ground and a wall in a rigid body simulator (left). From it, we construct an adversarial object (middle) that has identical surface geometries and first three moments of mass, and simulate it in a deformable simulator (middle). On the right, we show the difference in trajectories. Note that the angular difference means separation will grow over time and the effect can cascade through subsequent contacts.} 
 \label{fig:base-example}
\end{figure*}

\subsection{Dealing with the Constraints}
While the objective function in Eq. \ref{eq:discrete} is very simple, dealing with its constraints requires some care.

\subsubsection{Material Constraints}
The box constraints from Eq. \ref{eq:discrete}(a-c) are used to ensure that the adversarial object is made up of physically reasonable, perceptually rigid materials by setting an allowable range of Young's moduli, Poisson's ratio, and density.

While the material occupancy constraint from Eq \ref{eq:discrete}(d) is a discrete value of either 0 or 1, we follow the SIMP approach of treating it as a continuous variable between $[\alpha_\text{min}, 1]$ where the optimization drives it towards those endpoints, and then snapping it to the discrete values as a postprocessing step. This allows elements ranging from fully occupied to unoccupied, while not running into the simulation errors that would arise from 0 occupancy elements \cite{Mlejnek1992SomeAO}. To satisfy the boundary occupancy constraint from Eq. \ref{eq:discrete}(e), we simply set the values of $\alpha$ corresponding to boundary elements to $1$, and reduce the $\alpha$ degrees of freedom to the interior elements.

We satisfy these constraints by parameterizing the materials with new per element variables $\theta_{\{Y, \nu, \rho, \alpha\}}$:
\begin{equation}
\label{eq:parameterization}
\begin{split}
	x &= f_\text{param}(\theta, x_\text{min}, x_\text{max}) \\ &= \frac{1}{2} \tanh\left(\theta \right)(x_\text{max} - x_\text{min}) +  \frac{1}{2}(x_\text{max} + x_\text{min}),
\end{split}
\end{equation}
where $x$ is a stand-in for a material parameter, $\theta$ is its corresponding parameterized variable, and $x_\text{min}$ and $x_\text{max}$ are its minimum and maximum value. 
Now any value of $\theta$ will correspond to a material parameter in the allowable range, and thus implicitly satisfying the constraints while using an unconstrained optimization over $\theta$ (see Fig \ref{fig:didactic-example}). 

\subsubsection{Moments of Mass Constraints}
Finding a physically plausible occupancy and mass density distribution over the object that satisfies the mass moments constraints from Eq. \ref{eq:discrete}(f-h) is straightforward, and there are likely many such distributions for given target moments given the very large number of degrees of freedom compared to the number of constraints. 

We can compute the mass moments exactly via quadrature rule, using the per element volume vector $A$, mass densities $\rho = f_\text{param}(\theta_\rho)$, and material occupancies $\alpha = f_\text{param}(\theta_\alpha)$.  Since the mass densities and material occupancies are constant per element, this ultimately means that we can construct a moments of mass operator $S(V, T) \in \mathbb{R}^{10\ \times\ |T|}$  that acts on the per element (effective) density to compute the mass moments with the quadrature rule. Each row in $S$ corresponds to a different moment. Then, the adversarial object's moments are found as a matrix-vector multiplication:
\begin{equation}
\label{eq:moment-construction}
	M_{adv} = S (\alpha \odot \rho),
\end{equation}
where $\odot$ refers to the Hadamard product (and $\alpha \odot \rho$ is the effective density). 
We choose to enforce Eq. \ref{eq:discrete}(f-h) via soft constraint.

\subsection{Final Optimization Problem}
\begin{figure*}[t]
	\centering
	\includegraphics[width=\linewidth]{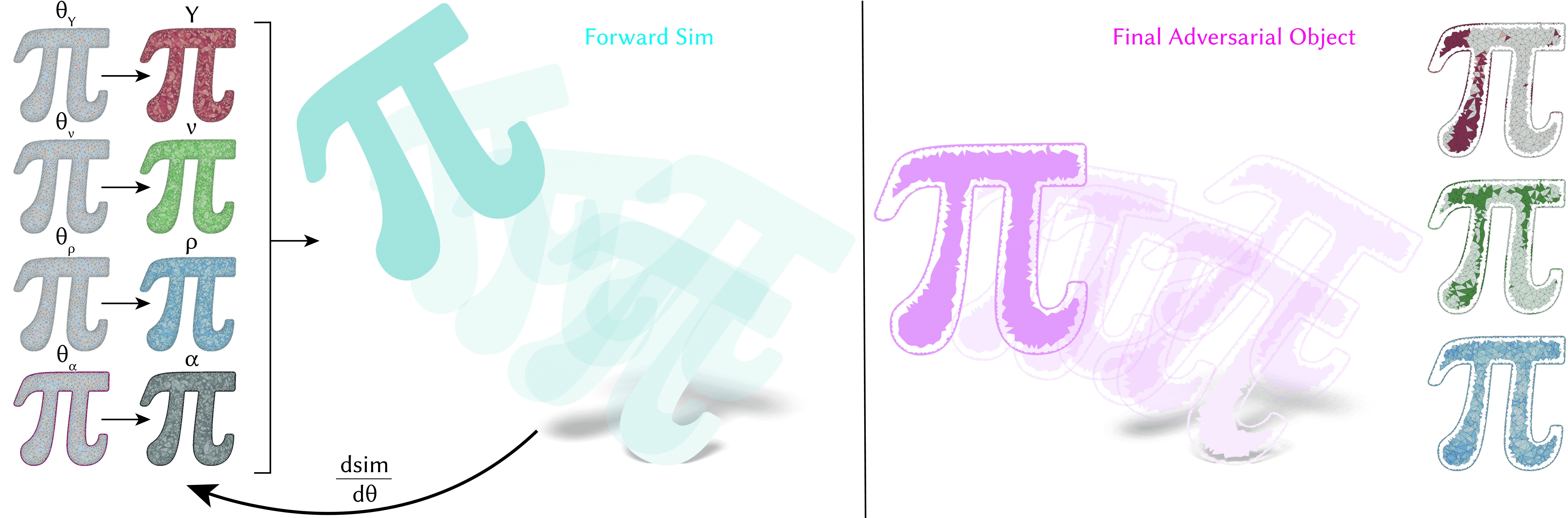}
	\vspace{-0.7cm}
	\caption{Our differentiable physics simulator allows us to construct adversarial objects using a first order optimization method.}
	\label{fig:didactic-example}
\end{figure*}
Using the parameterization scheme and soft constraint formulation for the moments of mass, we simplify the optimization problem in Eq. \ref{eq:discrete}:
\begin{equation}
\label{eq:total-cost}
\begin{split}
	 \underset{\theta_Y, \theta_\nu, \theta_\rho, \theta_\alpha}{\mathrm{argmin}} &\  -|| q_\text{adv}(t_\text{end}) - q_\text{ref}(t_\text{end}) ||_M^2\ + \\
	 \ &\  \beta \ || M_\text{ref} - M_\text{adv} ||^2,
\end{split}
\end{equation}
which we can be solved using a standard first-order descent method - we choose ADAM \cite{adam} (see Fig \ref{fig:didactic-example}). We determine $\beta$ experimentally. 
\revision{
While the use of soft constraint means the mass moments of the adversarial object are not guaranteed to exactly match those of the reference, appropriate choice of the coefficient $\beta$ results in very low error (see Table \ref{table:moments-comparison}), such that rigid body trajectory differences will be negligible on the timescales of interest.
}

\subsection{Forward Simulation}
Due to the increased stability and the ability to take larger timesteps, implicit integrators are commonly used for elastodynamics simulation. We use BDF-2 due to it's superior damping performance in comparison to implicit Euler \cite{bdf2-book}.

Using one of the standard techniques in simulation, we pose each dynamic step as the solution to a non-linear optimization problem where forces (including contact) are described as gradients of continuous potential energy functions (see e.g. \cite{ipc, implicit-newton}). Our hyperelastic constitutive model is the stable Neohookean energy from \citet{dynamic-deformables}, and our contact forces are from the smoothly clamped barrier energy in \citet{ipc}.
The ``energy'' in each time step is fully differentiable, but to compute the solution we may require many Newton iterations with line-search. For an explicit formulation, see Supplemental (A).

It is now clear how the material parameters are incorporated; the mass matrix $M$ must account for the densities and occupancies, and the hyperelastic strain energy $E_\Psi$ must account for the Young's modulus, Poisson's ratio, and occupancies. Following the topology optimization literature, we raise $\alpha$ to the power of 3 when it is used to scale the stiffnesses in $E_\Psi$ in order to promote occupancy sparsity \cite{topology-opt-reference}.

From the total energy of the system, we can can analytically calculate its gradient and Hessian (which we project to be positive definite), and run a Newton solve to convergence at each timestep until the end of the simulation.

\subsection{Simulation Gradients}
Our optimization problem is now well posed, but simply backpropagating through the nested loops over timestep and Newton iterations is untenable. 
Like much of the inverse design community, we instead leverage the Adjoint Method (summarized in \cite{mcnamara2004fluid}), which is used to differentiate through constrained optimization problems of the form:
\begin{equation}
\begin{split}
	\mathop{\text{argmin}}_\theta & \text{ } f(q, \theta) \\
	\text{s.t.} & \text{ }g(q, \theta) = 0,
\end{split}
\end{equation}
where $q$ is the state, $\theta$ is the design parameters to optimize, $f(q, \theta)$ is the cost function, $g(q,\theta)$ is the constraint. In inverse physics problems, $g$ is typically a constraint requiring that $q$ and $\theta$ are physically valid. 
Then, the gradient can be calculated as:
\begin{equation}
\label{eq:adjoint-equation}
\frac{d f}{d \theta} = \frac{\partial f}{\partial \theta} - \left(\frac{\partial f}{\partial q} \left(\frac{\partial g}{\partial q}\right)^{-1} \right) \frac{\partial g}{\partial \theta}.
\end{equation}

By using the Adjoint method to provide the derivatives for each timestep, we can avoid the massive performance hit of using auto-differentiation through Newton solves while still taking advantage of the convenience afforded by auto-differentiation frameworks \cite{gradsim}. The optimization problem from the forward simulation is implicitly in the form of the adjoint problem:
$$ q_{t+1} = argmin_q \ E(q, \theta) \ \text{ s.t. }\ G(q, \theta) = 0, $$ 
where $E$ is the minimization objective and $G$ its gradient ($\frac{\partial E}{\partial q_{t+1}}$). See Supplemental (B) for implementation details.

\subsubsection{Gradient Smoothing}
One of the limitations of nodal/element based topology optimization is the existence of the so-called non-physical ``checkerboard patterns'' where the optimized material occupancies are allocated such that patches are connected only at the element corners due to mesh/grid connectivity. One of the standard approaches to avoiding this is by using gradient smoothing, which is generally referred to in topology optimization as applying a ``filter function'' \cite{top-opt-filters}. In topology optimization, the design domain is usually discretized as a regular grid, and the filter functions are typically hat functions centered at each element. 

\begin{figure}[t]
	\includegraphics[width=\linewidth]{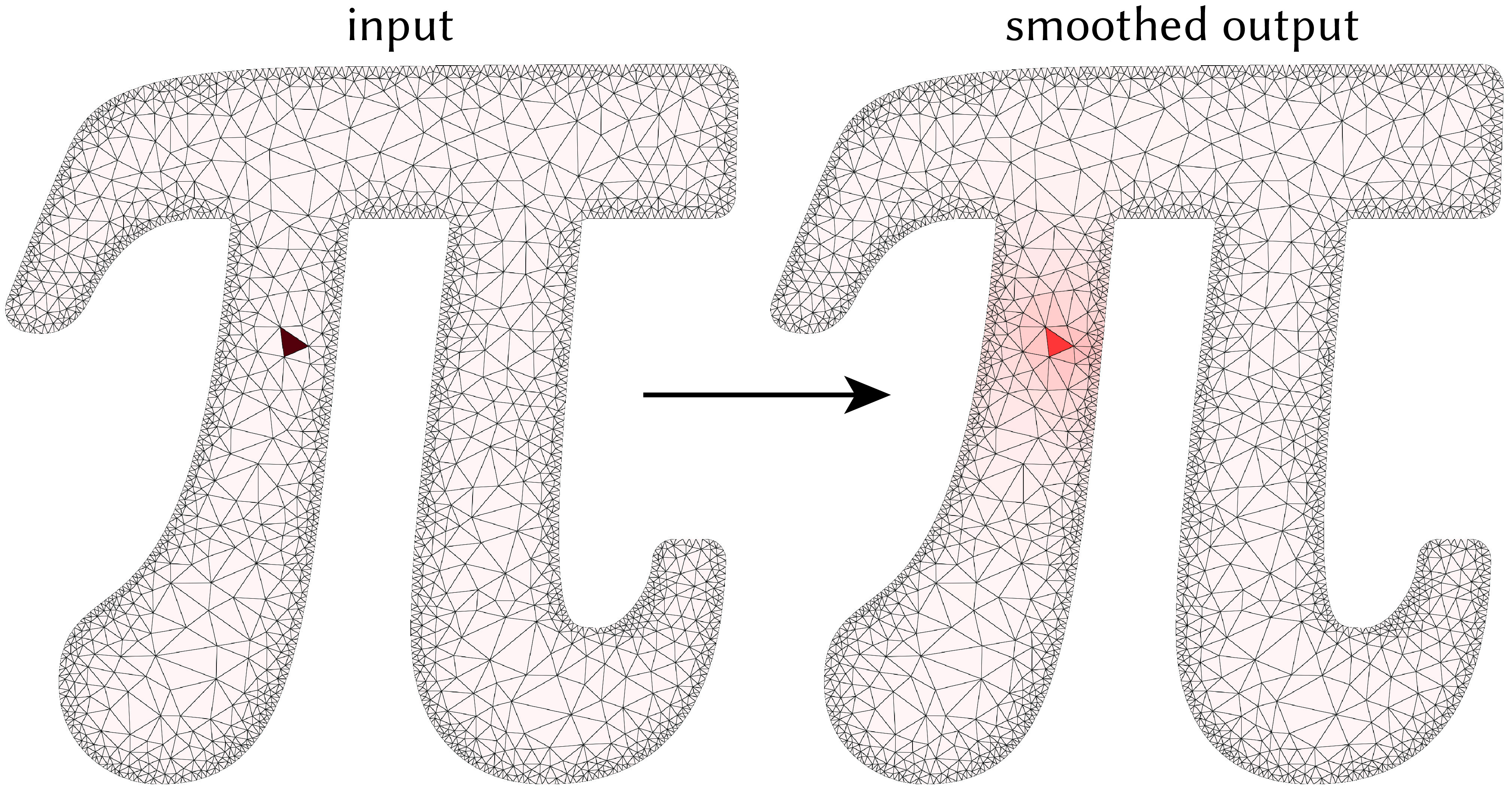}
	\vspace{-0.7cm}
	\caption{Our geometrically motivated smoothing operator from Eq. \ref{eq:gradient-blurring} is used to smooth an element defined function on an irregular mesh.}
	\label{fig:smooth-function}
\end{figure}

As we encounter irregular meshes, we instead use a more geometrically motivated mesh Laplacian based smoothing proposed by \citet{alec-blurring}. 
The standard cotangent Laplacian $L$ acts on vertex valued functions while our occupancy is per element. So, we instead first construct the false barycentric subdivision of the mesh so that the original elements have corresponding vertices. We can then apply Laplacian blurring using this subdivided mesh, and use a matrix $N = [0\ I]$, to transfer per element values to the barycenter vertices. Thus, we blur a per element input $\delta$ as:
\begin{equation}
\label{eq:gradient-blurring}
	\hat{\delta} = N^T (\tilde{M} - \gamma \tilde{L})^{-1} N A\  \delta, 
\end{equation}
where $\tilde{L}$ is the subdivided mesh Laplacian, and $\tilde{M}$ a modified mass matrix for the subdivided mesh, with the barycenter vertices having an area of the original mesh elements and all other vertices having area of 0 (see Fig. \ref{fig:smooth-function}). This smoothing operator is also conservative - it will preserve the volume weighted integral of the input function.

\begin{figure}[t]
	\includegraphics[width=\linewidth]{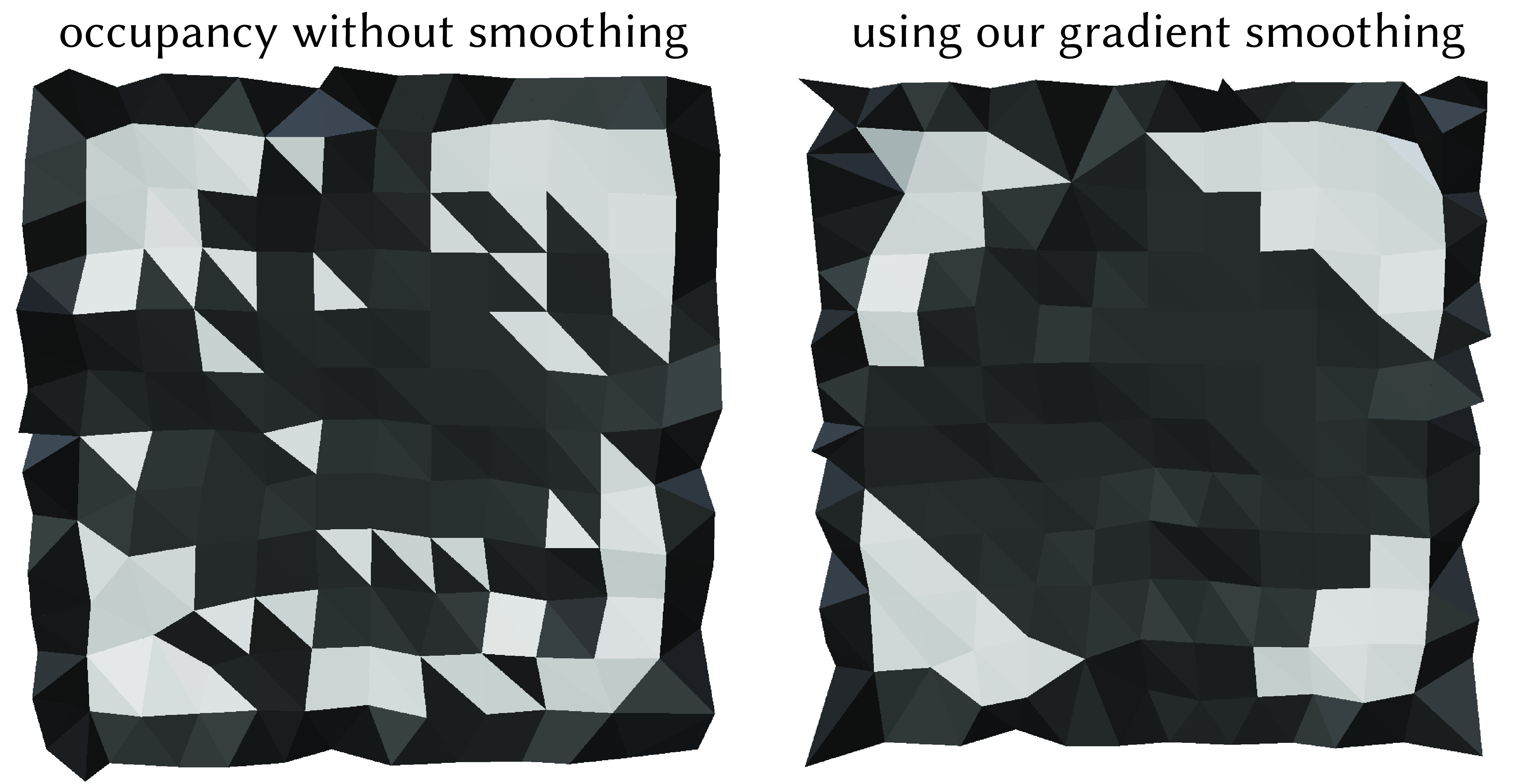}
	\vspace{-0.7cm}
	\caption{Using our geometrically motivated smoothing operator on the material occupancy gradients of this adversarial cube, we can avoid the checkerboarding artifacts that otherwise appear.}
	\label{fig:topology-opt-checkerboard}
\end{figure}

We apply this smoothing method on the gradient of the material occupancies ($\theta_\alpha$) to avoid the aforementioned checkerboarding artifacts (see Fig. \ref{fig:topology-opt-checkerboard}). The smoothing matrix from Eq. \ref{eq:gradient-blurring} is of size $| T | \times | T|$, while the $\theta_\alpha$ is of size $| T |_\text{interior}$, so we use selection matrices to apply it. 

\subsection{Postprocessing}
In SIMP, raising $\alpha$ to the power of 3 will drive the occupancies towards extremes by penalizing intermediary values - the effective volume is proportional to $\alpha$ but the effective stiffness is less than proportional (see \citet{topology-opt-reference} for a derivation and in depth discussion). 
However, this is not enough to guarantee no intermediate occupancy values. Therefore, as a post-processing step after the materials are determined, we sharpen the occupancies by rounding $\alpha$ so that its elements are either 0 or 1.
We then have an optimization to ensure the mass moments constraint is still satisfied:
\begin{equation}
	\underset{\theta_\rho}{\mathrm{argmin}} \text{    } || M_\text{ref} - M_\text{adv} ||^2,
\end{equation}
which we solve by running gradient descent starting with our previously found $\theta_\rho$.

\section{Implementation and Experiments}\label{sec:results}
\begin{figure*}[t]
 \centering
  \includegraphics[width=\linewidth]{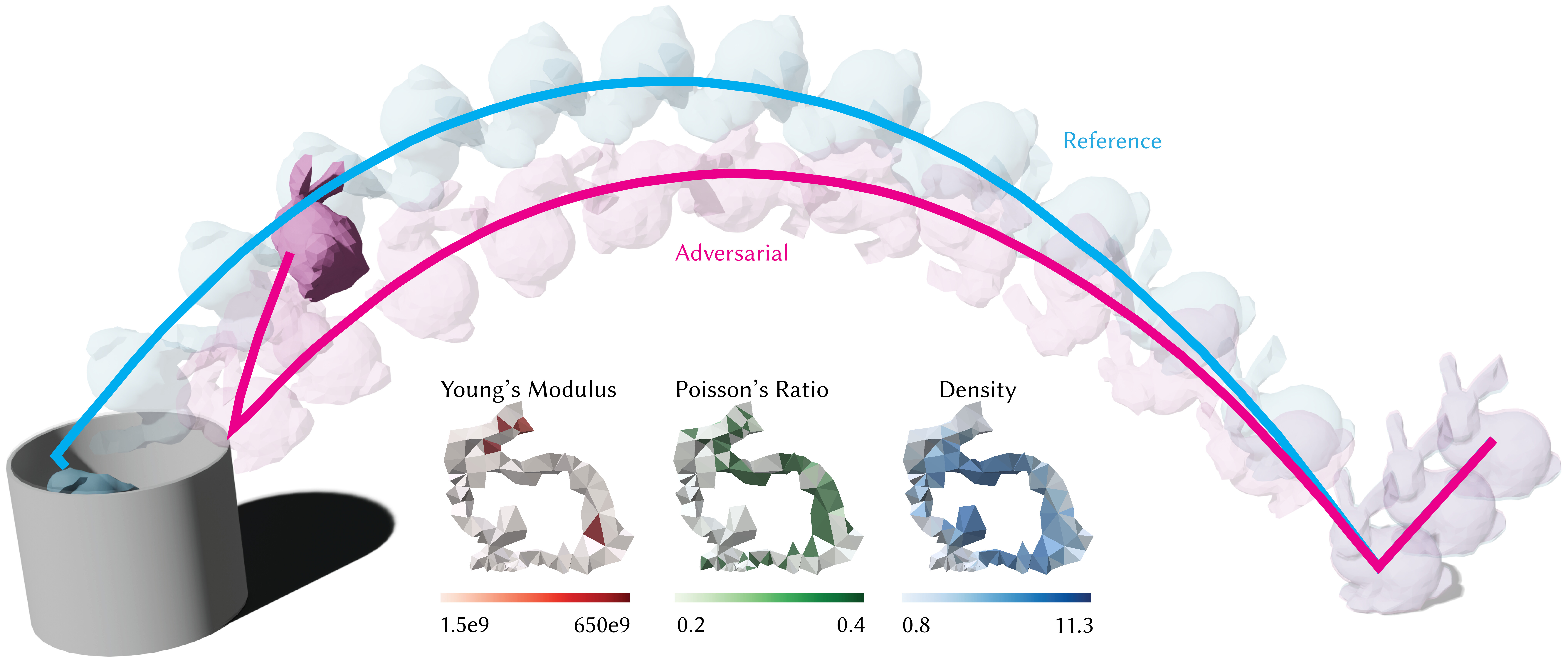} 
 \caption{Starting with a reference bunny (blue) which has a trajectory going into a bin, we construct an adversarial bunny (pink). A greater portion of the kinetic energy in the adversarial case is rotational, causing it to travel less distance and instead bounce off the rim of the bin.} 
 \label{fig:bunny-example}
\end{figure*}
\begin{table*}[t]
\caption{Comparison of reference and adversarial moments of mass for all of our examples. In all cases, the moments closely match.}
\label{table:moments-comparison}
\rowcolors{2}{white}{cvprblue!25}
\resizebox{\textwidth}{!}{%
\begin{tabular}{|lrrrrrrrrrr|}
    \hline
    \textit{Object} & \textbf{$m_0$} & \textbf{$m_{1,x}$} & \textbf{$m_{1,y}$} & \textbf{$m_{1,z}$} & \textbf{$m_{2,xx}$} & \textbf{$m_{2,yy}$} & \textbf{$m_{2,zz}$} & \textbf{$m_{2,xy}$}  & \textbf{$m_{2,xz}$} & \textbf{$m_{2,yz}$} \\
    Ball (ref) & 1.825e+01 & -3.215e-05 & -4.682e-05 & 3.930e-05 & 1.058e-01 & 1.059e-01 &  1.055e-01 &  7.475e-08 & -1.050e-06 &  2.079e-06 \\
    Ball (adv) & 1.825e+01 & -3.215e-05 & -4.682e-05 &  3.930e-05 &  1.058e-01 &1.059e-01 &  1.055e-01 &  7.462e-08 & -1.050e-06 &  2.080e-06 \\
    Star (ref) & 8.811e+02 &  3.314e-03 & -3.201e-03 &  1.016e-02 &  2.178e+02 &1.155e+02 & 1.155e+02 &-5.076e-04 & 1.560e-03 &-4.151e-03 \\
    Star (adv) & 8.811e+02 &  3.311e-03 & -3.200e-03 & 1.016e-02 & 2.178e+02 &1.155e+02 & 1.155e+02 & -5.259e-04 &  1.565e-03 & -4.142e-03 \\
    Bunny (ref) & 3.901e+03& -2.403e-13& 3.411e-13&-4.263e-13&1.382e+03&8.974e+02& 1.028e+03&9.705e-01&  4.553e+01& 2.682e+02 \\
    Bunny (adv) & 3.901e+03&-1.920e-07& 2.939e-08& 2.131e-06&  1.382e+03&8.974e+02&  1.028e+03& 9.705e-01& 4.553e+01&  2.682e+02 \\
    Cube (ref) & 3.124e-01 & 9.922e-07 &  3.564e-07 & 2.571e-07 & 1.301e-04 &1.301e-04 & 1.301e-04 & 1.164e-08 & 3.443e-08 &-1.833e-08 \\
    Cube (adv) & 3.124e-01&  9.922e-07&  3.565e-07&  2.571e-07& 1.301e-04&1.301e-04& 1.301e-04&  1.164e-08& 3.443e-08& -1.833e-08 \\
    Bat (ref) & 4.132e+00 & 3.082e-05 & -2.229e-05 & 2.287e+00 & 1.453e+00&1.453e+00&  1.713e-03& 2.258e-07& -1.887e-05&  2.489e-05 \\
    Bat(adv) & 4.132e+00 &-8.139e-05 & -4.273e-05 &  2.287e+00& 1.453e+00&1.453e+00&  1.930e-03& 3.406e-06& -2.030e-05&  2.410e-05 \\
    \hline
\end{tabular}}
\end{table*}

We implement our simulation in \textsc{Python}, using \textsc{PyTorch}  \cite{pytorch-citation} as our autodifferentiation framework. We use the sparse direct solver \textsc{Cholespy} \cite{inverse-rendering-geometry} for all of our linear system solves. We use \textsc{libigl} \cite{libigl} for all necessary geometry processing, and \textsc{Blender} and \textsc{Polyscope} \cite{polyscope} for visualization.
Our simulations are run with a timestep of 0.01 seconds, and an IPC barrier distance of $10^{-3}$ unless stated otherwise. We run all of our experiments on CPU on a 2021 M1 Macbook Pro. We demonstrate the power of our method by conducting black box attacks, where our rigid reference objects are simulated in \textsc{Bullet} \cite{pybullet}, and adversarial objects are constructed using our custom simulator and demonstrated on \textsc{Polyfem} \cite{polyfem}. 

Unless noted otherwise, we use following bounds on material parameters for all examples. The Young's modulus is allowed to range from 2.5 GPa to 650 GPa which corresponds respectively to ABS plastic and tungsten carbide. The Poisson's ratio is allowed to range from 0.2 to 0.4 which corresponds to most metals and plastics. The mass density is allowed to range from 0.8 g/cc to 11.3 g/cc, which corresponds respectively to PMP plastic and lead. 

We find that due to the chaotic nature of contact, it is possible to successfully create adversarial results even with very simple object and contact geometries. 
In Fig. \ref{fig:teaser}, we construct an adversarial ball from just the first contact off the backboard, and the resulting trajectory difference is greatly amplified by the subsequent contacts off the rim. Similarly, in Fig. \ref{fig:bunny-example}, we construct an adversarial bunny from the first collision off the ground, such that it has a different center of mass trajectory than its reference, with more energy being stored in rotational kinetic energy. This trajectory difference is sufficient for the bunny to bounce off the rim of a bin rather than fall into it.
In Fig. \ref{fig:base-example}, we have a more complex geometry, with a star bouncing off of two perpendicular planes. In this case, our optimized optimization includes contact off of both planes. The star example is run with a timestep of 1/30 seconds, and the minimum Young's modulus is raised to 5GPa. Due to the thin shape of the star, we only optimize the Young's modulus, Poisson's ratio, and density; we leave the material occupancy fixed.

\begin{figure}[t]
	\includegraphics[width=\linewidth]{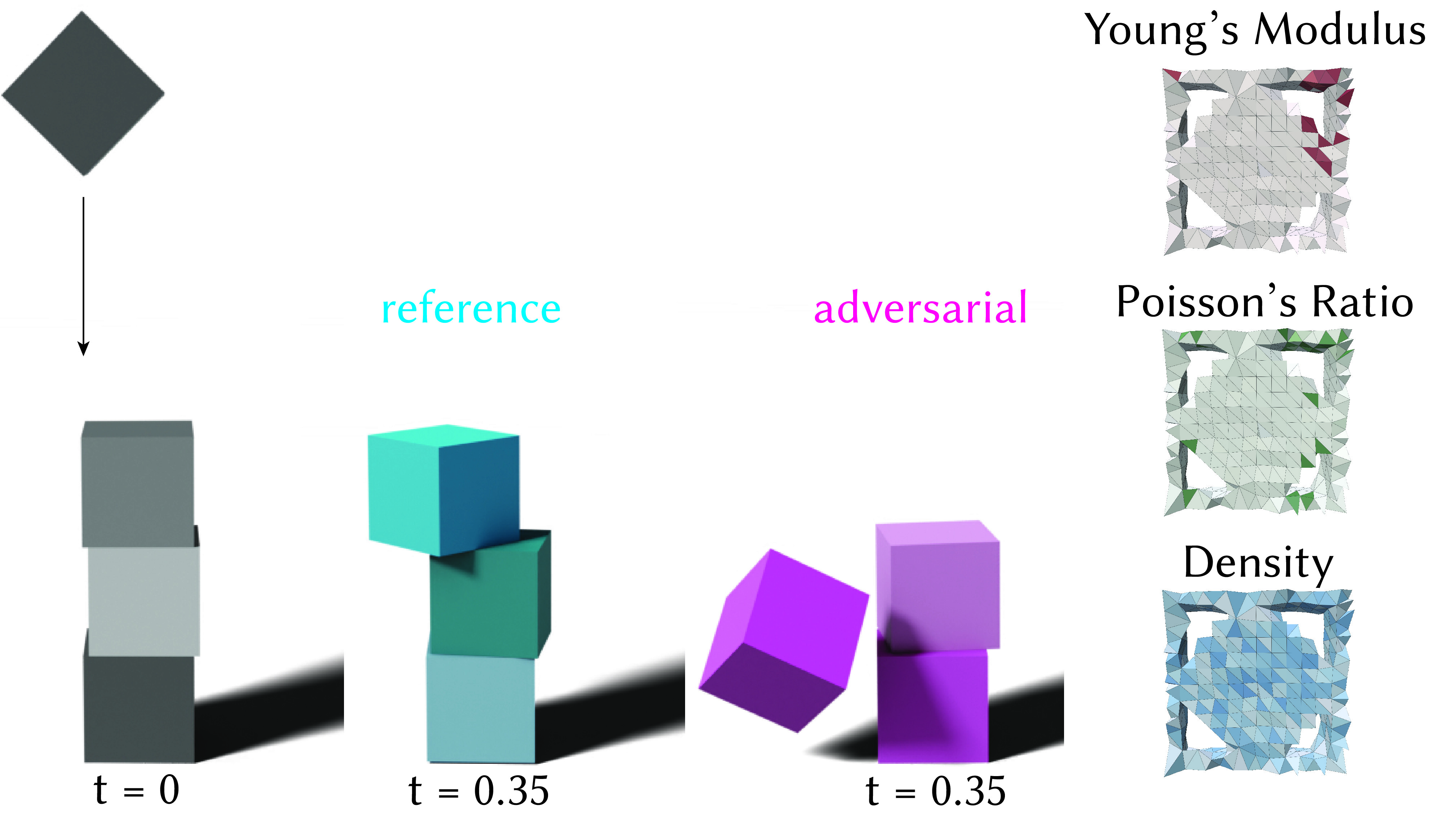}
	\vspace{-0.7cm}
 \caption{Here, we simulate a collision between a block and a stack of blocks. In the reference blocks (left), the stack is left intact after the collision. Using adversarial blocks (right), the stack falls down.} 
	 \label{fig:stacking}
\end{figure}

In Fig. \ref{fig:stacking}, we show a more complex scenario with a cube striking a stack of three other cubes with a timestep of 1e-3 seconds. Using reference blocks, the stack remains upright post collision, while with adversarial blocks, it falls over. The adversarial cube is created from planar impacts from above and below. Even though our deformable simulation is frictionless, these adversarial cubes are successful at attacking a simulator with a friction coefficient of 0.4.

\begin{figure}[t]
	\includegraphics[width=\linewidth]{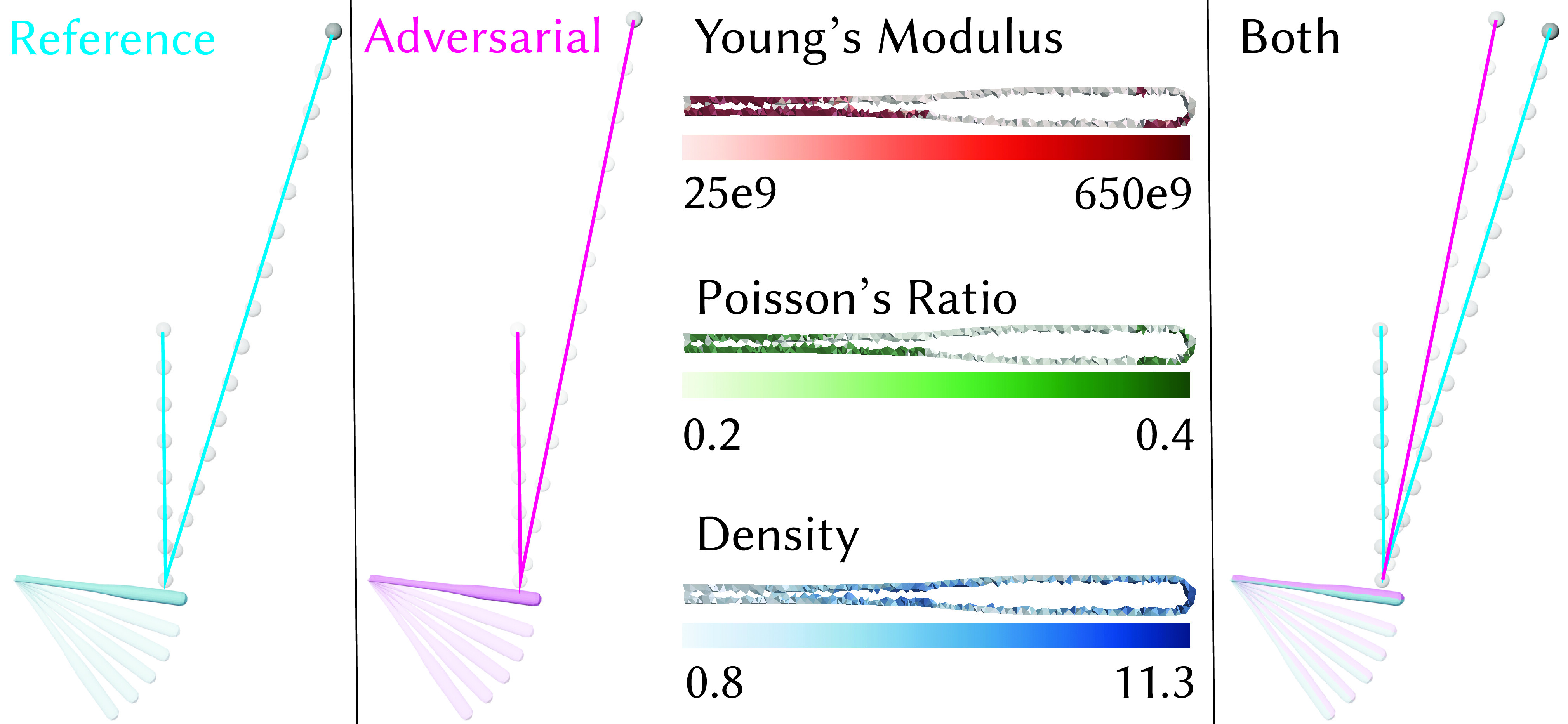}
	\vspace{-0.7cm}
 \caption{We construct an adversarial bat that aims to get the trajectory of the struck ball as close as possible towards the center (i.e. $x=0$). } 
	 \label{fig:baseball-example}
\end{figure}

Thus far, all examples have been undirected attacks. In Fig. \ref{fig:baseball-example}, we change our cost function to demonstrate a directed attack, where an adversarial bat is constructed specifically so that the trajectory of the struck ball is as close to the center as possible causing a significant angular deviation from the reference.

For all of these examples, the comparison of their moments of mass and that of the corresponding reference object is given in Table \ref{table:moments-comparison}. For all reference objects, we determine moments by using a constant mass density of 2.5g/cc. We provide further experimental details in the Supplemental, as well as baseline comparisons with simulations of stiff uniform material objects.
\section{Conclusion and Future Work} \label{sec:conclusion}
In this paper we introduce the adversarial setting of rigid body simulators, and present a method of constructing perceptually rigid adversarial objects that will behave identically to a reference object in a rigid body simulation but maximally different in deformable simulations. We do so by implementing a differentiable deformable simulator, and defining a cost function that encodes both the difference in the trajectories of the adversarial and reference object as well as the first three moments of mass. We then minimize this cost using the ADAM algorithm. We have shown over several examples that there can be significant deviations in the trajectories even through simple contact. 
\revision{
While it is unknown how common these adversarial objects can accidentally occur or how hard physical adversarial objects would be to manufacture, knowing that they can happen at all is interesting and important in consideration of rigid body simulation tools being used in safety critical applications (e.g. robotics control in industrial settings).
}

The primary limitation of our method is that the materials parameters are allowed to continuously span some range. This means that even though the material parameters individually are in reasonable ranges, there is no guarantee that a constructed material will actually exist. In the future, it would be interesting to incorporate a materials library that can be populated with real world materials. By our choice of hyperelastic constitutive model, we are limited to isotropic materials. One direction of future work could be to investigate if we can cause even greater trajectory differences by allowing anisotropic materials or multiscale modelling, where even microstructures are considered in the optimization. Another limitation is that our method can occasionally result in disconnected pieces after rounding occupancies, which we currently fix as a postprocess (via \texttt{steiner\_tree} \cite{networkx}). We leave more sophisticated topological postprocesses (e.g., \cite{joining-cv}) or reparameterizations (e.g., \cite{fast-quasi-harmonic}) as future work. Our method currently considers a fully deterministic setting and a single trajectory. It would be interesting to study the construction and performance of adversarial objects in a stochastic setting.

\revision{
Having identified a deficiency in current simulation techniques, it is interesting to consider how they can be improved. An immediate first approach could be using our adversarial objects to improve robustness of learned physics models through adversarial training.  
More broadly, stiff dynamics simulation is typically seen as a dichotomy between fast rigid body simulation, and very slow deformable simulation. In computer graphics and animation, there has been research in between those regimes, for example, elastodynamics modelling via adding visual effects to augment rigid body simulation (e.g. \cite{comp-dynamics}).
Our work suggests there could be value in studying more accurate methods for stiff materials. Recent work in this area includes \citet{bounce-map}, who incorporate spatially varying coefficient of restitution to rigid body simulation. Interesting future work in simulation could be to incorporate our adversarial objects to evaluate methods in this area of research with the ultimate goal of sufficiently performant and accurate simulation techniques for stiff objects that can be used in the current robotics and ML pipelines.
}

We hope that our work encourages members of the machine learning and robotics community to investigate whether their rigid body simulations are sufficiently accurate for their needs and further consider what steps can be taken to improve simulator robustness.

{
    \small
    \bibliographystyle{ieeenat_fullname}
    \bibliography{references.bib}
}
\clearpage
\setcounter{page}{1}
\maketitlesupplementary
\appendix

In this supplemental document, we provide an explicit formulation of the optimization problem solved in each timestep of the forward simulation, an overview of calculating its derivatives with the adjoint method, as well as further details regarding experiments.

\section{Explicit Formulation of Forward Simulation Problem}\label{app:forward-sim}
To determine the optimization problem in the forward simulation, let us start with Newton's law: $F = Ma$. 
The BDF-2 scheme is given by: 
\begin{equation}
3x_{n+1} - 4x_n + x_{n-1} = 2h f(t_{n+1}, x_{n+1}).
\end{equation}

Writing Newton's law with the implicit formulation after time discretization gives us:
$$ M a_{n+1} = F_{n+1}.$$ 
Applying BDF-2 and substituting the forces with their corresponding differentiable potential energies, this expands to:
\begin{equation*}
\frac{1}{2h}M (3v_{n+1} - 4v_{n} + v_{n-1}) = -\left.\dv{\Phi}{q} \right\vert_{q_{n+1}}
\end{equation*}

Now, use the BDF-2 formula on the velocities to obtain an equation in terms of positions:
\begin{multline*}
\frac{1}{4h^2} M (3(3q_{n+1} - 4q_n + q_{n-1}) - 4(3q_n - 4q_{n-1} + q_{n-2}) + \\ (3q_{n-1} - 4q{n-2} + q_{n-3})) = -\left.\dv{\Phi}{q} \right\vert_{q_{n+1}}.
\end{multline*}
Simplifying, we get
\begin{multline*}
9 M (q_{n+1} - \frac{1}{9}(24q_n - 22q_{n-1} + 8q_{n-2} - q_{n-3})) =\\ -4h^2 \left.\dv{\Phi}{q} \right\vert_{q_{n+1}}.
\end{multline*}
Solving the forward step is then a root-finding problem, where we find the $q_{n+1}$ that is a zero of:
\begin{multline*}
f(q_{n+1}) = 9 M (q_{n+1} - \frac{1}{9}(24q_n - 22q_{n-1} + 8q_{n-2} - q_{n-3}))\ + \\4h^2 \left.\dv{\Phi}{q} \right\vert_{q_{n+1}}.
\end{multline*}

Let us introduce $\hat{q} = \frac{1}{9}(24q_n - 22q_{n-1} + 8q_{n-2} - q_{n-3})$, so that we can write:
\begin{equation*}
f = 9 M(q_{n+1} - \hat{q}) + 4h^2 \left.\dv{\Phi}{q} \right\vert_{q_{n+1}}.
\end{equation*}

Now, we can construct an energy $E$ such that $f = \dv{E}{q_{n+1}}$, so finding the root of $f$ is equivalent to minimizing the energy:
\begin{equation*}
E = \frac{9}{2} (q_{n+1} - \hat{q})^T M (q_{n+1} - \hat{q}) + 4 h^2 \Phi + c,
\end{equation*}
which is in turn equivalent to minimizing:
\begin{equation}
E = \frac{1}{2} (q_{n+1} - \hat{q})^T M (q_{n+1} - \hat{q}) + \frac{4}{9} h^2 \Phi.
\end{equation}

Splitting the potential energy into its constituent components yields the following optimization problem to find the positions at the end of the timestep:
\begin{multline}
\label{eq:sim-energy}
	q_{t+1} = \underset{q}{\mathrm{argmin}} \ \frac{1}{2} (q-\hat{q})^T M (q-\hat{q})\ + \\ \frac{4}{9}h^2 (\Phi_\Psi(q) + \Phi_g(q) + \Phi_c(q)),
\end{multline}
where $M$ is the FEM consistent mass matrix (see Eq. 12.55 of \citet{fem-book-rao} for tetrahedral definition), $\Phi_\Psi$ is the strain energy, $\Phi_g$ is the gravitational potential energy, and $\Phi_c$ is the contact potential energy.

For $ \Phi_\Psi$, we use the stable Neohookean energy density given in Eq 6.9 of \citet{dynamic-deformables}: 
\begin{equation}
\begin{split}
 \Phi_\Psi &= \int \Psi_{SNH}\ dV, \\
\Psi_{SNH} &= \frac{\mu}{2}(I_2-3) - \mu(I_3-1) + \frac{\lambda}{2}(I_3-1)^2,
\end{split}
\end{equation}
where $\lambda, \mu$ are the first two Lam\'e parameters (functions of $Y$, $\nu$). Note that when calculating this energy density, we must scale the stiffnesses by the cube of our material occupancy $\alpha$.

The gravitational potential energy is defined as: 
\begin{equation}
 \Phi_g = g^T M q.
\end{equation}

For the contact, we use the smoothly clamped barrier energy from \citet{ipc}:
\begin{equation}
   \Phi_c = \begin{cases} 
      -(d-\hat{d})^2 \log{\left(\frac{d}{\hat{d}}\right)} & 0 < d < \hat{d} \\
      0 & d \geq \hat{d}
   \end{cases}.
\end{equation}

\section{Simulation Derivatives}\label{app:adjoint-derivs}
Since we are using reverse mode auto-differentiation, we are given the gradient of the final cost $C$ with respect to the simulation step output (typically called \textsc{grad\_output} in Torch literature).
Thus, using the definition of the adjoint method (Eq. 7 from the main document), we find:
\begin{equation}
\begin{split}
	\frac{\partial C }{\partial \hat{q}} \bigg\rvert_{q_{t+1}}&= (\textsc{grad\_output}^T H^{-1}) M \\
	\frac{\partial C}{\partial \theta_{\{Y, \nu\}}} \bigg\rvert_{q_{t+1}}&= -\frac{4}{9}h^2 (\textsc{grad\_output}^T H^{-1}) \frac{\partial^2 E_\Psi}{\partial q \partial \theta_{\{Y, \nu\}}} \\
	\frac{\partial C}{\partial \theta_{\{\rho, \alpha\}}} \bigg\rvert_{q_{t+1}}&= - (\textsc{grad\_output}^T H^{-1}) \frac{\partial}{\partial \theta_{\{\rho, \alpha\}}} M (q\ - \\ \hat{q} + \frac{4}{9}h^2 g),
\end{split}
\end{equation}
where we already have the per-timestep energy Hessian $H$ ($\frac{\partial^2 E}{\partial q_{t+1}^2}$) from the forward simulation, and we can analytically compute the Jacobians. 

\section{Experiment Details}
In this section we provide additional details regarding the setup for the experiments we have conducted. Unless specified otherwise, we use a timestep of 0.01s, an IPC barrier distance of 1e-3 m, gravitational acceleration of -9.8 m/s${}^2$, a Young's modulus range from 2.5 GPa to 650 GPa, a Poisson's ratio range from 0.2 to 0.4, and a mass density range from 0.8 g/cc to 11.3 g/cc. In our examples, we use ADAM optimization parameters of $\beta_1 = 0.7$ and $\beta_2 = 0.95$. We run all simulations to convergence.

\subsection{Adversarial Ball}
The adversarial ball scenario is constructed to mimic a standard basketball setup. The ball is constructed from a tetrahedral mesh of a sphere with radius 0.121m and is comprised of 1820 vertices and 9056 tetrahedra. The backboard assembly is constructed out of a plane and a rim, following NBA dimensions. The backboard assembly is placed at position [0, 0, 3.048] m, and is positionally constrained (treated as a collision mesh rather than simulated). The ball is placed at position  [3.4, -3.287, 1.8495] m, and is released with velocity [-3.516,  3.516,  6.099] m/s. 

While the ball collides off the planar backboard ($xz$ plane) and has multiple collisions off of the rim, we construct the adversarial ball using just the first impact off the backboard. To do this, we first simulate the trajectory of the ball from the release point described above, and capture its state just prior to the collision off the backboard. We then use this state as the initial condition for  the 0.25s simulations of the ball bouncing off a plane positioned to match the backboard that we use to construct the adversarial object. The optimized adversarial ball is demonstrated from the initial conditions detailed above for a longer 1.5s simulation. In the black box attack demonstration, we observe a very large qualitative difference in the trajectories - in reference case, the ball successfully goes into the hoop and in the adversarial case, it bounces off the rim and out.

\subsection{Adversarial Star}
In the adversarial star example, we launch a star off of two perpendicular planes (located at $z = 0$ and $y = 0$). The star is of radius 1m, and is comprised of 440 vertices and 1506 tetrahedra. The star has an initial position of [0, 2.5, 2.0] m, and a velocity of  [0, -10, -10] m/s. In this example, we disable gravity and use a timestep of $\frac{1}{30}$s. Due to the thin profile of the star, we choose to optimize only the Young's modulus, Poisson's ratio, and mass density (i.e. we keep the star as a solid object, abstaining from any topology optimization). Additionally, we increase the minimum Young's modulus to allow a range of 5GPa to 650GPa.
To construct the adversarial example we simulate 0.3s, but the demonstration of the attack against the simulation uses a longer 1s simulation. In this experiment, we observe a roughly 8 degree angular separation between the reference and adversarial trajectories.

\subsection{Adversarial Bunny}
In the adversarial bunny example, a bunny is bounced off of the ground towards a bin. Similarly to the adversarial ball, the adversarial bunny is constructed using just the first planar contact. The bunny mesh consists of 699 vertices and 2274 tetrahedra. To construct the adversarial example, the bunny is given an initial position of  [0, 1.5, 1.5] m, and an initial velocity of  [0, -7.75, -10] m/s. It is simulated for 0.2s during which it bounces off of the $xy$ plane. 
To demonstrate the attack, the bunny has an initial position of  [0, 2.0, 2.0] m, an initial velocity of  [0, -7.75, -10] m/s, and the bin is placed at [0, -13.75, 0] m. These simulations are run for 2.5s. Similarly to the adversarial ball example, we have a major qualitative difference in the simulation result, with the reference bunny successfully going into the bin and the adversarial bunny failing to do so.

\subsection{Adversarial Cubes}
The adversarial cubes example differs from the previous experiments as there are multiple bodies being simulated, there is non-zero friction, and the simulation used to construct the adversarial example differs greatly from the simulation used to evaluate it. To construct the adversarial cube, we use a simplified setup where the cube of side length 5cm is placed at [0, 0, 0.05] m, is given an initial velocity of [0, 0, -0.1] m/s. The cube bounces off the $xy$ plane, and again off a plane above it at $z = 0.1$. The cube mesh contains 2167 vertices and 10164 tetrahedra.

The evaluation of this example involves three identical cubes stacked (offset) atop each other, and a fourth cube strikes the stack from above (inspired by Figure 6 of \citet{backward-rigid-body}). The three cubes are placed at: [0, 0, 0.026]m,  [0, -0.0225, 0.65]m, and  [0, 0, 0.128]m. The fourth cube is rotated 45 degrees in the x-axis, and is given an initial position of [0, -0.0225, 0.65] m and an initial velocity of [0, 0, -3.0] m/s. These simulations use a timestep of 1e-3s and a friction coefficient of 0.4. In this experiment, the stack of reference cubes successfully stays upright post collision, and the top cube in the stack of adversarial cubes falls over.

\subsection{Adversarial Bat}
The adversarial bat example contains two different simulated bodies - the bat which contains the degrees of freedom, and the ball whose trajectory is optimized. This case also differs from the previous examples in that it is a directed attack rather than undirected. The optimization objective is to get the ball as close as possible to the center (i.e. $x= 0$). This gives us an optimization cost term of $ || q_\text{adv}(t_\text{end})_x ||^2 $. The bat consists of 2214 vertices and 9098 tetrahedra. The swing of the bat is encoded by choosing the 14 vertices on the base of the bat to be used as Dirichlet boundary conditions; at each timestep of the simulation, the energy minimization problem has a constraint that pins the boundary condition vertices to their positions corresponding to the swing. We solve this using the standard extension to Newton's method for feasible start nonlinear optimization problems with equality constraints. In this way, we give the bat a constant angular velocity of $1.25 \pi$ rad/s, and placed with its base at the origin with an initial orientation of -55 degrees about the $z$ axis. The ball is given an initial position of  [0.75, 1.25, 0.245] m, with an initial velocity of  [0,  -6,  0] m/s. To construct the adversarial bat, we follow the same procedure as with the adversarial ball example, running the simulation forward to capture the state shortly before contact, and using that state as the initial conditions for the simulation in the optimization loop. For constructing the adversarial example, the simulation duration is 0.25 s, and for demonstrating it is 2.0 s. As with the star, we increase the minimum Young's modulus to 25 GPa to keep the bat's motion perceptually rigid. In this experiment, we observe a roughly 5 degree angular difference between the rigid body and adversarial trajectories.

\section{Baseline Simulations}
Different simulation tools work very differently under the hood, for instance, the use of barrier functions vs linear complementarity problem for contact modelling, choice of integrator, etc. 
One might wonder to what extent the trajectory difference of the adversarial object is due to the construction of the object itself as opposed to the differences in the simulation tools used. We investigate these concerns by running baseline simulations in our deformable simulator (\textsc{Polyfem}) with extremely stiff, uniform material objects.

Ultimately, both rigid body and deformable simulators are meant to model real world phenomena. Thus, by taking measures such as choosing appropriate simulation parameters and using highly stiff materials, one would expect that the simulated results match across simulators. 
For the baseline material parameters, we use a Young's modulus of 1e13 Pa, a Poisson's ratio of 0.28, and a mass density of 2.5 g/cc.
We choose to run the baseline simulation using the same timesteps and contact parameters used in the general experiments.

\begin{figure*}[t]
 \centering
  \includegraphics[width=\linewidth]{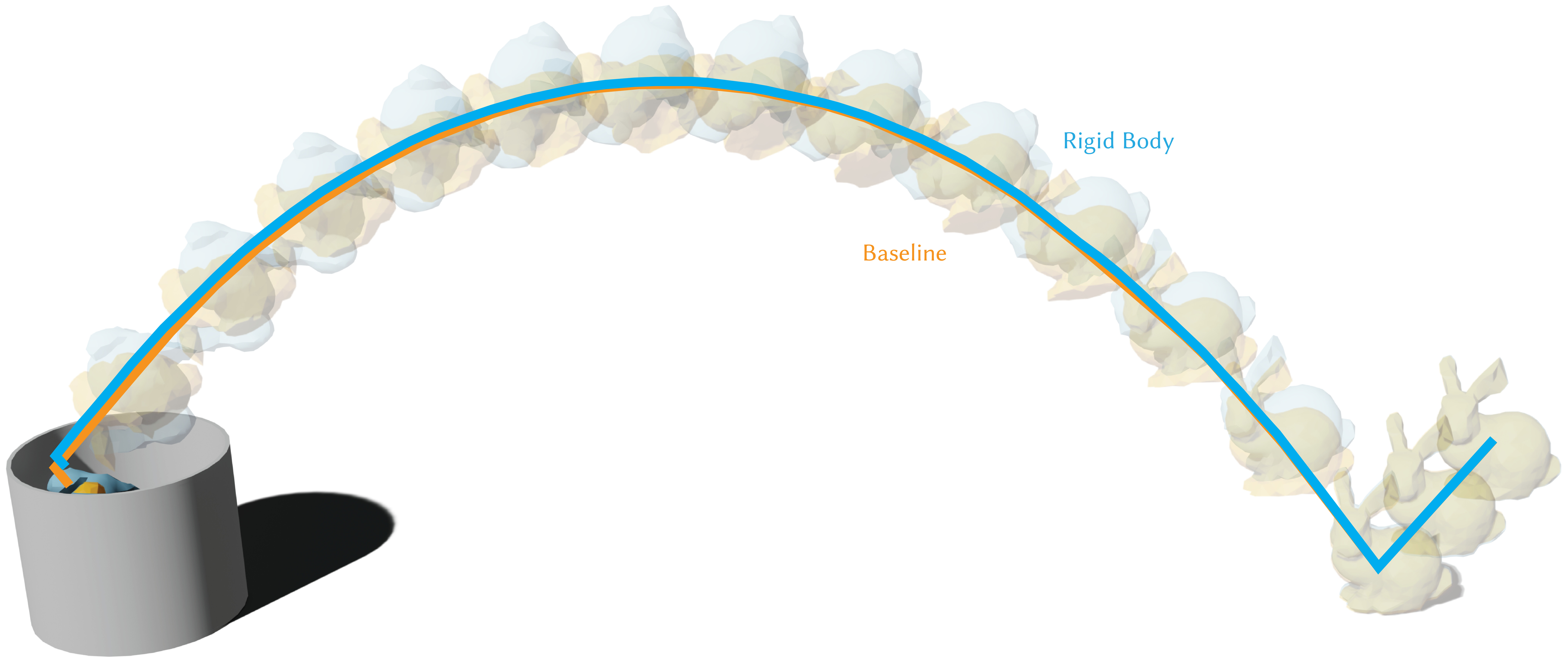}
 \caption{Following the setup from Fig. \ref{fig:bunny-example}, the baseline simulation (gold) matches the rigid body simulation's (blue) defining characteristic of successfully going into the bin. Note that the trajectories are in very close agreement.} 
 \label{fig:bunny-baseline}
\end{figure*}

\begin{figure}[t]
	\includegraphics[width=\linewidth]{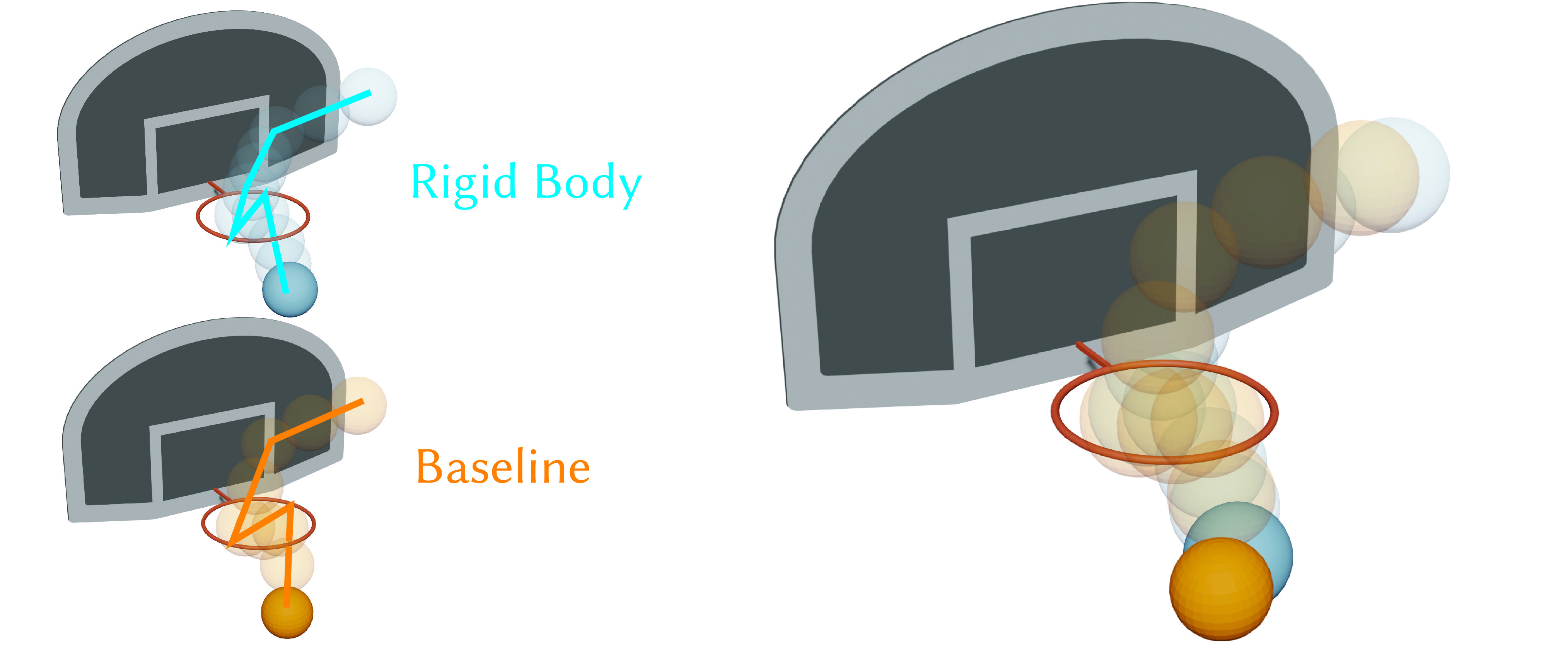}
	\vspace{-0.7cm}
 \caption{Following the setup from Fig. \ref{fig:teaser}, the baseline simulation (gold) matches the rigid body simulation's (blue) defining characteristic of successfully going into the hoop.} 
	 \label{fig:basketball-baseline}
\end{figure}

\begin{figure}[t]
	\includegraphics[width=\linewidth]{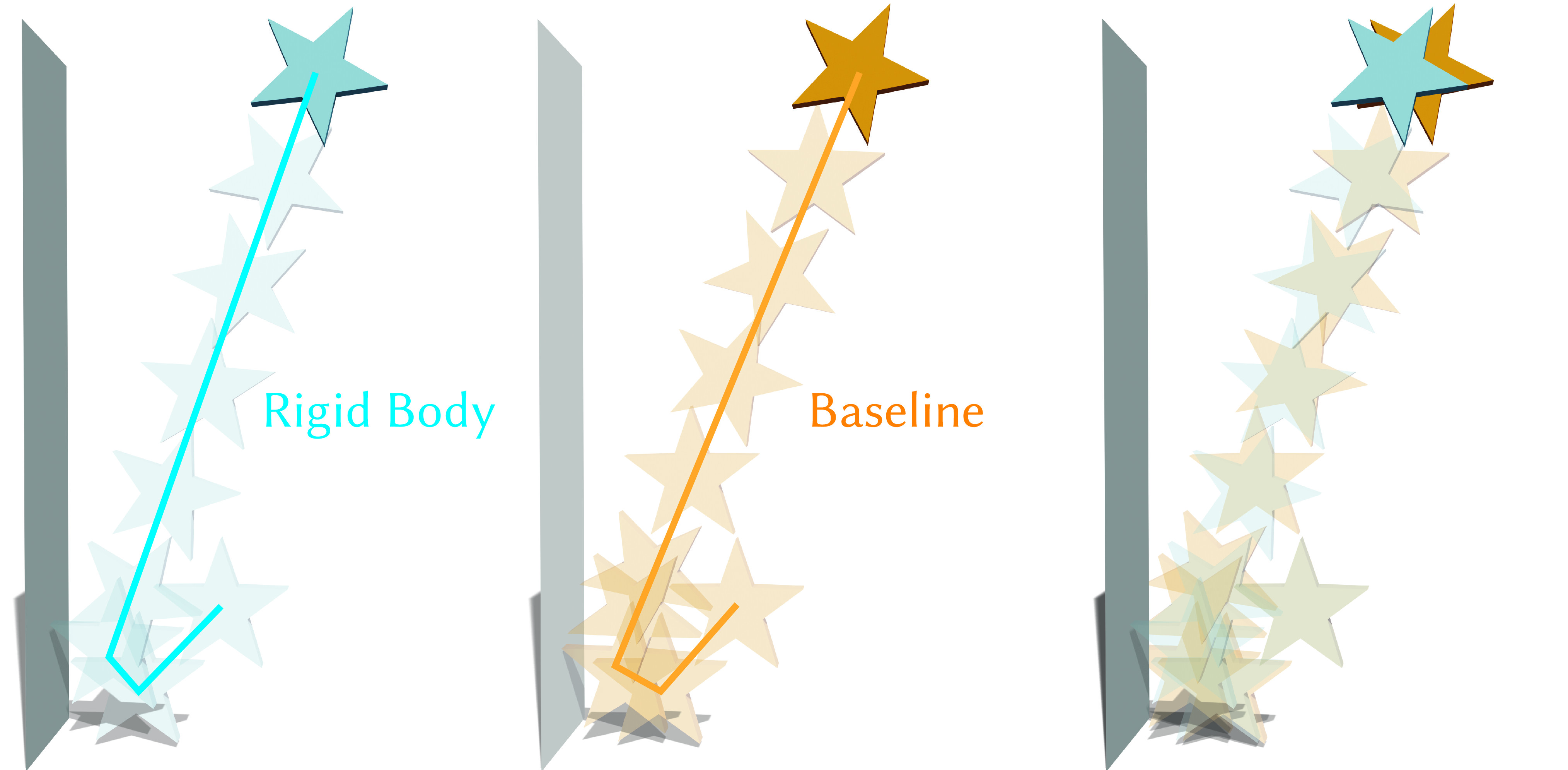}
	\vspace{-0.7cm}
 \caption{The trajectory of the baseline simulation (gold) corresponding to the star example from Fig \ref{fig:base-example} is close to that of the rigid body simulation (blue), with a roughly 2 degree angular difference.} 
	 \label{fig:star-baseline}
\end{figure}

\begin{figure}[t]
	\includegraphics[width=\linewidth]{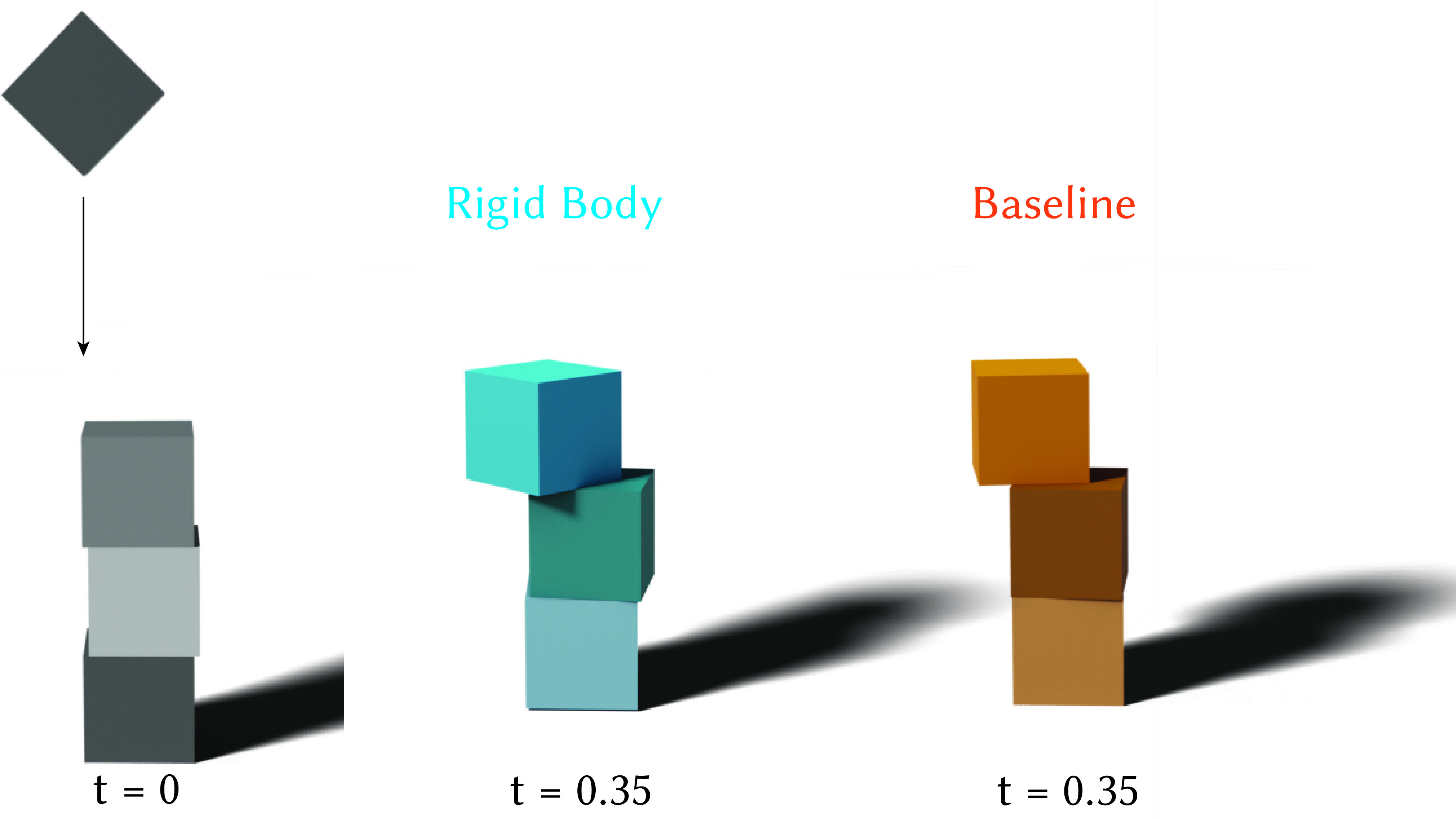}
	\vspace{-0.7cm}
 \caption{Here, we simulate a collision between a block and a stack of blocks using a stiff deformable simulation (gold). The post collision state looks similar to that of the corresponding rigid body simulation (blue), with the stacks left intact after the collision.} 
	 \label{fig:stacking-baseline}
\end{figure}

\begin{figure}[t]
	\includegraphics[width=\linewidth]{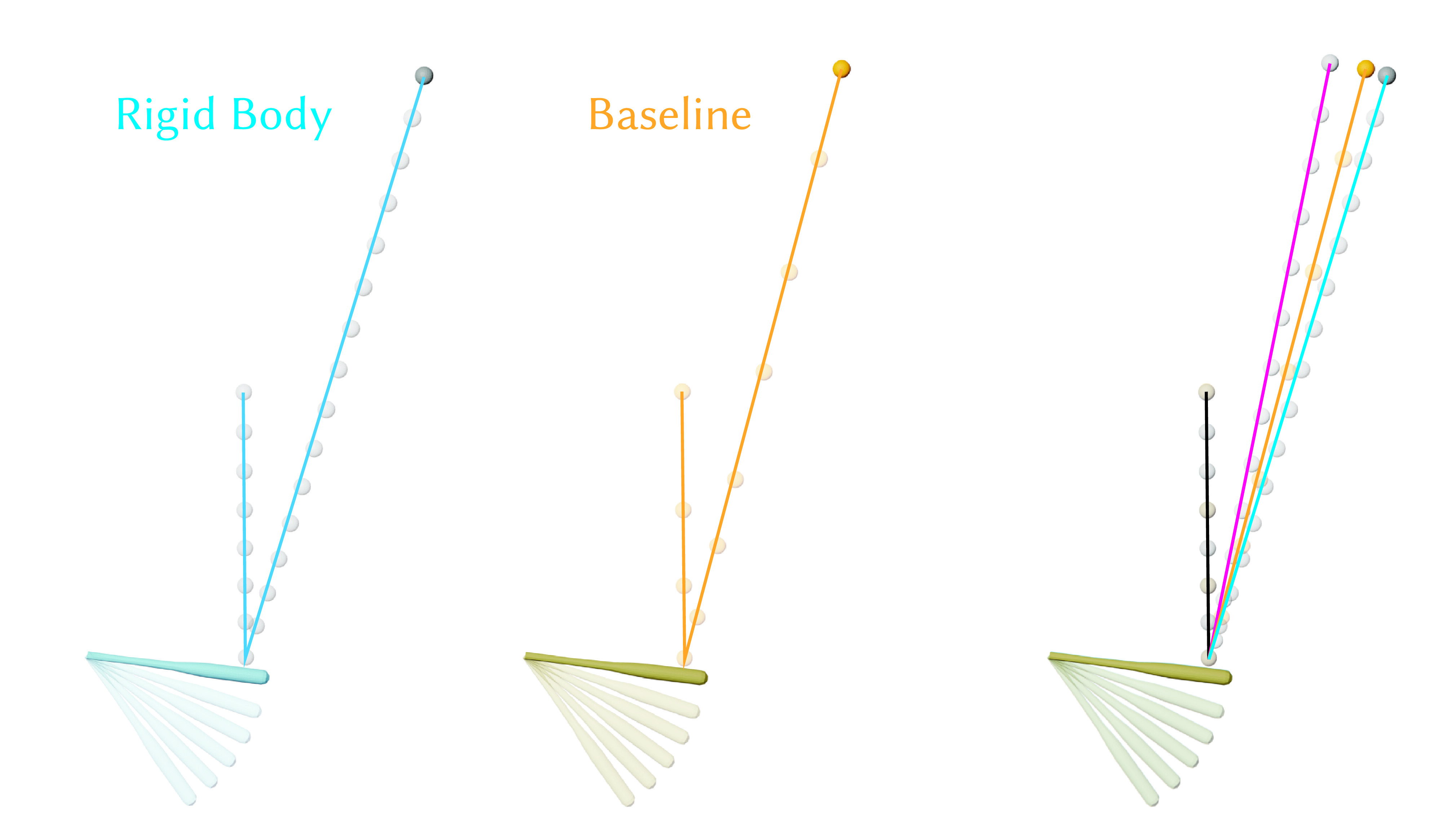}
	\vspace{-0.7cm}
 \caption{The baseline simulation (gold) of the bat example from Fig. \ref{fig:baseball-example} shows poorer agreement than the other examples. We see a roughly 2 degree angular difference between the rigid body (blue) and baseline trajectories.}
	 \label{fig:baseball-baseline}
\end{figure}

The results of these baseline simulations are show in figures \ref{fig:bunny-baseline}, \ref{fig:basketball-baseline}, \ref{fig:star-baseline}, \ref{fig:stacking-baseline}, and \ref{fig:baseball-baseline}. We see that the baseline simulations match the rigid body simulations reasonably well. In the three experiments that have a qualitative "success / failure" outcomes (adversarial ball, bunny, cubes), the baseline simulations match the rigid body simulations. For the star and bat examples, the baseline simulation trajectories is about two degrees off from the rigid body reference. While this difference is certainly nontrivial, we note that in both cases the baseline simulation is closer to the rigid body reference than it is to the adversarial examples.

We believe the main sources of the inaccuracies between the simulators are our large timesteps, the internal mesh contact sampling, and the use of single value coefficient of restitution in the rigid body simulators (standard in rigid body simulation). It may be possible to get much better agreement by decreasing the timestep, exploring different levels of mesh refinement, and using more advanced restitution models such as citet{bounce-maps-restitution}.

\end{document}